\documentclass{article}
\usepackage[T1]{fontenc}
\usepackage{iclr2027_conference,times}
\usepackage[hyphens]{url}
\usepackage{graphicx}
\usepackage{booktabs}
\usepackage{amsmath,amssymb}
\usepackage{placeins}
\usepackage{float}
\usepackage{adjustbox}
\usepackage{xurl}
\usepackage{hyperref}
\hypersetup{hidelinks,pdftitle={Source-Lifted Flow Matching for Intervenable Multimodal Imitation},pdfauthor={He Zhang, Ying Sun, Ziyang Chen, Qicheng Luo, Yiren Zhao, Weiyu Guo, Pengteng Li, Yandong Guo, Hui Xiong}}
\graphicspath{{./}}
\newcommand{\methodfull}{Source-Lifted Flow Matching}
\newcommand{\method}{SL-FM}
\newcommand{\mech}{Orthogonal Source Lifting}

\newcounter{algorithm}
\title{Source-Lifted Flow Matching\\for Intervenable Multimodal Imitation}
\author{\textbf{He Zhang, Ying Sun, Ziyang Chen, Qicheng Luo, Yiren Zhao,}\\
\textbf{Weiyu Guo, Pengteng Li, Yandong Guo, Hui Xiong}}
\iclrfinalcopy

\begin{document}
\maketitle
\lhead{}

\begin{abstract}

Flow-matching policies are promising for imitation learning because they model complex multimodal action distributions. However, their stochasticity is largely passive: repeated sampling may yield diverse behaviors, but users cannot directly choose among valid continuations from the same state. We propose \textbf{Source-Lifted Flow Matching (SL-FM)}, a source-intervenable flow-matching policy that exposes such a handle while keeping the velocity field shared and latent-free (without a separate discrete-handle input). The handle selects only the source endpoint of the conditional flow, not a mode-specific field, preserving the standard formulation while avoiding decomposition into separate mode-conditioned dynamics.
The core mechanism is \textbf{Orthogonal Source Lifting}, designed to prevent path-crossing ambiguity. Instead of partitioning target actions by mode, SL-FM lifts handle-specific sources into auxiliary orthogonal coordinates and keeps targets in the original action subspace. This preserves the demonstrated action distribution while allowing one shared field to carry different branches without merging at crossings. To keep handles usable across states, we learn a state-dependent source mixture end to end and use a responsibility floor, giving each handle weak supervision and mitigating dead modes.
Experiments on crossing-flow diagnostics and robot-control benchmarks show that SL-FM converts passive source randomness into an actionable intervention variable. It removes crossing-induced composite trajectories, changes future routes in 91.1\% of matched-prefix interventions, and achieves strong free-deployment performance, with improvements in several benchmark settings. Overall, source geometry provides actionable multimodal control without conditioning the velocity field on the selected mode.

\end{abstract}

\section{Introduction}
\suppressfloats[t]

\begin{figure}[t]
\centering
\includegraphics[width=0.92\linewidth]{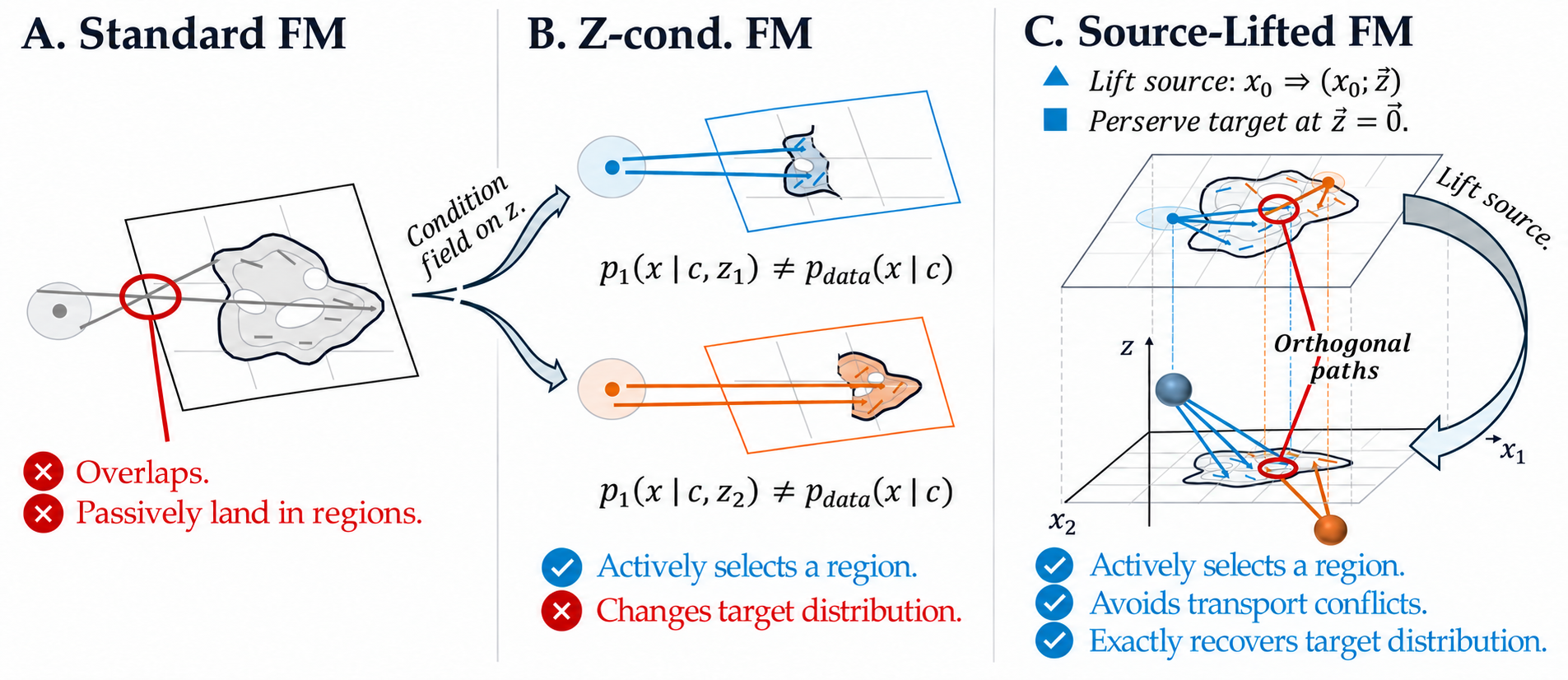}
\caption{Main idea of Source-Lifted Flow Matching. 
Standard flow matching passively samples multimodal behaviors, while mode-conditioned fields may split both the target distribution and velocity field. 
SL-FM instead lifts handle-specific sources into orthogonal coordinates, keeps all targets in the original action subspace, and uses one shared latent-free field to preserve source identity through crossings.}
\label{fig:main-concept}
\end{figure}

Recent advances in imitation learning have increasingly leveraged generative policies to model complex, multimodal action distributions~\citep{chi2023diffusionpolicy,pearce2023imitating,jiang2025streamingflow}.
Unlike conventional behavioral cloning methods that often assume a unimodal Gaussian policy~\citep{florence2022ibc,shafiullah2022bet}, diffusion and flow-matching policies can represent multiple plausible actions for the same observation~\citep{chi2023diffusionpolicy,lipman2023flowmatching,jiang2025streamingflow}, which is essential in robot tasks with several valid strategies~\citep{jia2024d3il,chi2023diffusionpolicy}. For example, in obstacle avoidance, demonstrations may pass an obstacle from either side; a good policy should preserve both alternatives rather than collapse them into an averaged action.

However, representing diverse behaviors is not the same as controlling them. Existing flow-matching policies usually use source noise as a passive sampling mechanism~\citep{lipman2023flowmatching,jiang2025streamingflow}: resampling the source may produce different rollouts, but the user cannot directly choose which continuation will be realized from a fixed decision state. This distinction matters for downstream planning, human-in-the-loop control, and robot tasks where several valid futures share the same prefix. A policy that sometimes goes left and sometimes goes right is useful; one that lets a planner choose the continuation from the same local state is more useful.

A straightforward way to expose such a choice is to condition the policy or the velocity field on a discrete mode variable~\citep{guo2025variationalrfm,zhai2025vfp}. While effective, this changes the nature of the model. A velocity field of the form $v_\theta(x,t,s,z)$ can resolve multimodality by learning separate mode-specific dynamics, making the discrete input an expert selector rather than an intervention on the source of a single shared flow. Our goal is different: we ask whether a conditional flow-matching policy can retain one shared velocity field while exposing a source handle that can be selected at test time.

To this end, we introduce \textbf{Source-Lifted Flow Matching (SL-FM)}, a source-intervenable flow-matching policy for multimodal imitation learning. SL-FM assigns each state a small set of source handles. In free deployment, the handle is sampled from a learned state-dependent prior. Under intervention, it can be set externally at a decision state. Crucially, the handle only determines where the conditional flow starts. The velocity field receives the intermediate point, time, and observation, but never the discrete handle itself. This preserves the standard conditional flow-matching formulation and forces behavior selection to arise from source geometry rather than from mode-conditioned subfields (Figure~\ref{fig:main-concept}).

The key technical challenge is that source-only handles are not automatically reliable. If two multimodal paths cross in action space, a shared field may lose the identity of the chosen source branch and average incompatible velocities near the crossing. We address this problem with \textbf{Orthogonal Source Lifting}. Instead of partitioning the demonstrated action distribution into $p(a \mid s,z)$, SL-FM lifts handle-specific sources into auxiliary orthogonal coordinates while keeping every target action in the original action subspace. The target action distribution is therefore unchanged, but different handles remain separated during transport, allowing one latent-free field to carry different branches without losing identity at crossings. We further learn a state-dependent source mixture end to end and introduce a responsibility floor, which gives every handle weak transport supervision and mitigates dead modes under redundant handle counts.

Our experiments evaluate both ordinary imitation performance and the proposed intervention interface. On a crossing-flow diagnostic, SL-FM removes composite trajectories caused by path crossings and enables direct branch selection through the source handle. On D3IL Avoiding, same-prefix counterfactual interventions show that changing only the local source handle redirects future behavior in 91.1\% of matched pairs. Across multimodal robot-control benchmarks, including D3IL and PushT, SL-FM maintains strong free-deployment performance and achieves the best average score among same-harness mechanism-matched baselines. These results show that source geometry can turn passive flow-matching randomness into an actionable multimodal control interface without conditioning the velocity field on the selected mode.

We summarize our \textbf{contributions} as follows:
\begin{itemize}
    \item We formulate source-intervenable conditional flow matching for imitation learning, where a discrete handle selects the source endpoint but never conditions the velocity field.
    \item We identify path-crossing identity collapse as a failure mode of source-only intervention in shared flow fields and introduce Orthogonal Source Lifting to preserve source identity through crossings.
    \item We learn a state-dependent source mixture with a responsibility floor to keep redundant handles trainable and mitigate dead modes.
    \item We evaluate SL-FM using both free-deployment benchmarks and same-prefix interventions, demonstrating an actionable source interface with strong imitation performance.
\end{itemize}

\section{Related Work}

\paragraph{Generative policies for multimodal imitation.}
Diffusion and flow-based policies have become strong generative models for multimodal imitation learning. Diffusion Policy, flow-matching policies, Streaming Flow Policy, and related implicit, tokenized, or score-based policies represent diverse action modes beyond unimodal Gaussian cloning~\citep{chi2023diffusionpolicy,lipman2023flowmatching,jiang2025streamingflow,florence2022ibc,shafiullah2022bet,reuss2023beso,pearce2023imitating}. Benchmarks such as D3IL and PushT further highlight the need to preserve multiple valid robot strategies rather than collapse them into averaged actions~\citep{jia2024d3il,chi2023diffusionpolicy}. This trend also appears in VLA and generalist robot policies, where systems such as $\pi_0$, $\pi_{0.5}$, and GR00T N1 use diffusion or flow-based action heads for continuous control~\citep{black2025pi0,physicalintelligence2025pi05,nvidia2025grootn1}. These works establish generative action modeling as a strong tool for marginal coverage; SL-FM instead asks whether the source randomness can be made locally intervenable under a matched rollout prefix.

\paragraph{Policy steering and latent conditioning.}
Another line of work exposes control by steering, adapting, or conditioning generative policies. Inference-time methods such as DynaGuide guide pretrained diffusion policies with external dynamics or task objectives~\citep{du2025dynaguide}. Adaptation and distillation methods, including DSRL, ReinFlow, GoldenStart, RFS, and related latent policy steering approaches, optimize noise variables, priors, entropy, residual actions, or latent actors for policy improvement~\citep{wagenmaker2025dsrl,zhang2025reinflow,zhang2026goldenstart,su2026rfs,ren2024dppo,park2025fql}. Variational Flow-Matching Policy is closer to our setting, using latent priors and mode-aware decoding for multimodal manipulation~\citep{zhai2025vfp}. These methods are complementary, but they expose control through guidance, value optimization, latent adaptation, or mode-conditioned generation. SL-FM targets a stricter interface: the handle only chooses where the conditional flow starts, while the velocity field remains shared and latent-free.

\paragraph{Source and coupling design in generative flows.}
The source distribution and source--target coupling strongly shape flow-matching geometry. OT-CFM, minibatch OT, rectified or optimal flow methods, CPD, MM-FM, and modal coupling design priors or pairings to shorten transport, reduce crossings, or improve sample efficiency, mostly in vision, generic, or motion generation~\citep{tong2024otcfm,pooladian2023multisample,liu2023rectifiedflow,albergo2023stochasticinterpolants,issachar2025cpd,luo2026mmfm,wang2025modalcoupling}. Control tasks add sequential decisions, state-dependent branching, and closed-loop feedback, so better transport geometry alone does not necessarily yield an actionable decision variable. SL-FM uses source structure for intervention rather than generation quality: it keeps the demonstrated target distribution intact, places the selectable handle only at the source, and tests whether changing that handle causally changes future behavior under a matched prefix.

\section{Preliminaries}

\paragraph{Notation.}
We consider an imitation dataset $\mathcal{D}=\{(s_i,a_i)\}_{i=1}^{N}$, where $s\in\mathcal{S}$ denotes the policy observation and $a\in\mathbb{R}^{d_a}$ denotes a demonstrated action or a flattened action chunk. Here $d_a$ is the action dimension, $t\in[0,1]$ denotes the artificial flow-matching time and the timesteps in tasks are indexed by $h$. The goal of imitation learning is to sample actions from the conditional demonstration distribution $p_{\mathrm{data}}(a\mid s)$ while executing successfully in closed loop.

\paragraph{Conditional flow matching for actions.}
Conditional flow matching learns a velocity field $v_\theta(x_t,t,s)$ that transports a source endpoint $x_0\sim p_0(\cdot\mid s)$ to a target endpoint $x_1=a\sim p_{\mathrm{data}}(\cdot\mid s)$~\citep{lipman2023flowmatching,tong2024otcfm}. For the linear conditional path
\begin{equation}
    x_t=(1-t)x_0+t x_1,\qquad 
    u_t=x_1-x_0,
\end{equation}
the standard regression objective is:
\begin{equation}
    \mathcal{L}_{\mathrm{CFM}}
    =
    \mathbb{E}_{(s,a)\sim\mathcal{D},\,x_0\sim p_0,\,t\sim \mathcal{U}[0,1]}
    \left[
    \left\|v_\theta(x_t,t,s)-u_t\right\|^2
    \right].
\end{equation}
At deployment, one samples $x_0$ from the source and integrates the ODE:
\begin{equation}
    \frac{d x_t}{dt}=v_\theta(x_t,t,s)
\end{equation}
from $t=0$ to $t=1$. The results are then executed as actions.

When $p_{\mathrm{data}}(a\mid s)$ is multimodal, a standard flow-matching policy can still produce diverse behaviors because different regions of the source space are transported to different action modes~\citep{lipman2023flowmatching,jiang2025streamingflow}. This is sufficient for marginal multimodal sampling, but the induced partition of the source space is implicit~\citep{jiang2025streamingflow}. Resampling the source may change the rollout, yet the user or planner is not given an explicit local variable that selects which continuation should occur.

\paragraph{From passive sampling to source intervention.}
Figure~\ref{fig:combined-diagnostics}(a) illustrates the distinction. At a decision state, multiple demonstrated futures may share the same prefix and branch into different valid continuations. A standard flow-matching policy may sample either branch, but this choice is passive: it is determined by source noise rather than by an externally selectable handle~\citep{jiang2025streamingflow}. We instead seek a source-intervenable policy with a discrete handle $z\in\{0,\ldots,K-1\}$ satisfying three requirements.
First, in free deployment, sampling $z$ should recover ordinary stochastic imitation behavior and preserve the target action marginal. 
Second, under intervention, a planner should be able to set $z=k$ at a local decision state while keeping the prefix fixed. 
Third, the velocity field must remain shared and latent-free, i.e., the model uses $v_\theta(\bar{x_t},t,s)$ rather than $v_\theta(x_t,t,s,z)$.

More formally, consider two rollouts that share the same environment seed, state-action prefix, and stochastic draws before a decision step $h$. A source intervention changes only the local handle, from $z=k$ to $z=k'$, and then resumes the same policy interface. Let $\rho(\cdot)$ denote an evaluation-time route or continuation label, not a training label. The desired behavior is that changing the handle can change the future continuation,
\begin{equation}
    \rho(\xi^{k}_{h:}) \neq \rho(\xi^{k'}_{h:}),
\end{equation}
while both branches remain valid task executions. This is stronger than showing diversity across independent rollouts: it asks whether the policy exposes a causal local variable for choosing among futures from the same prefix.

This problem setup motivates the construction in the next section. Since the handle is not allowed to condition the velocity field directly, any intervention effect must be carried by the geometry of the source. The remaining challenge is to make that source identity survive multimodal path crossings, where a shared field may otherwise average incompatible branch velocities.

\FloatBarrier
\section{Methodology}
\label{sec:methodology}

\begin{figure}[t]
\centering
\includegraphics[width=0.98\textwidth]{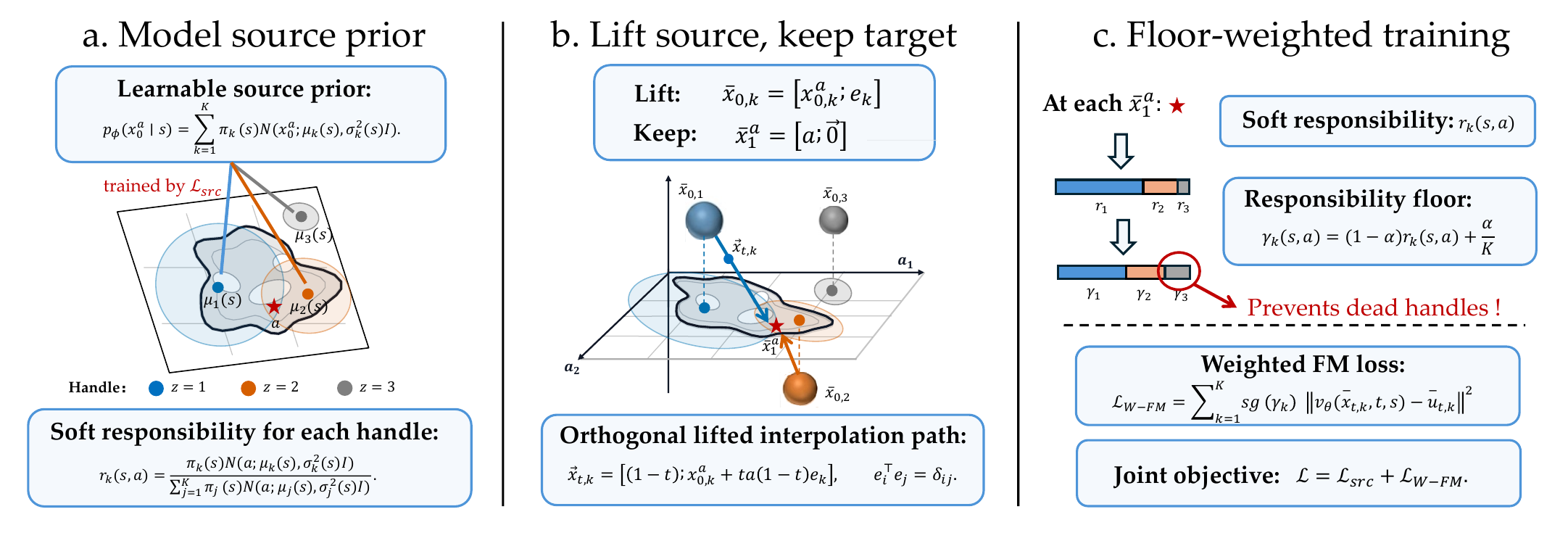}
\caption{
Method overview of Source-Lifted Flow Matching. 
SL-FM models a state-conditioned source prior, lifts handle-specific sources into auxiliary orthogonal coordinates while keeping all targets on the zero-lift plane, and trains one shared latent-free velocity field with floor-weighted responsibilities. 
For visual clarity, the schematic shows fixed anchors and omits scalar loss weights.
}
\label{fig:method-source-lift}
\end{figure}

The previous section defines the desired interface: a discrete source handle should be selectable at a local decision state, while the velocity field itself remains shared and not explicitly conditioned on that handle. We now introduce \textbf{Source-Lifted Flow Matching (SL-FM)}, which realizes this interface through three components, as illustrated in Figure~\ref{fig:method-source-lift}: a state-conditioned source prior, Orthogonal Source Lifting, and floor-weighted flow training.

\subsection{Modeling the Source Prior}

Standard flow matching starts generation from an unstructured source, such as a standard Gaussian. SL-FM instead models the source as a small state-conditioned mixture, where each mixture component corresponds to a selectable source handle~\citep{bishop1994mdn,jacobs1991moe}.
Let $a\in\mathbb{R}^{d_a}$ be the demonstrated action, and let $x_0^a\in\mathbb{R}^{d_a}$ denote the action-space source point before lifting. We use $z\in\{0,\ldots,K-1\}$ to index $K$ source handles. For each state $s$, a prior network predicts the parameters of an isotropic Gaussian mixture:
\[
    \{\pi_k(s),\mu_k(s),\sigma_k(s)\}_{k=0}^{K-1},
\]
where $\pi_k(s)$ are mixture weights with $\sum_k \pi_k(s)=1$, $\mu_k(s)\in\mathbb{R}^{d_a}$ are component means, $\sigma_k(s)>0$ are diagonal scales, and $I_{d_a}$ is the $d_a$-dimensional identity matrix. The resulting source prior is:
\begin{equation}
    p_\phi(x_0^a\mid s)
    =
    \sum_{k=0}^{K-1}
    \pi_k(s)
    \mathcal{N}\!\left(x_0^a;\mu_k(s),\sigma_k^2(s)I_{d_a}\right).
    \label{eq:source-prior}
\end{equation}
Free deployment samples $z\sim\pi_\phi(\cdot\mid s)$ and then samples from the selected Gaussian component,
\begin{equation}
    x_{0,k}^{a}=\mu_k(s)+\sigma_k(s)\epsilon_k,
    \qquad
    \epsilon_k\sim\mathcal{N}(0,I_{d_a}).
    \label{eq:source-sample}
\end{equation}
Under intervention, the evaluator directly sets $z=k$ and uses the corresponding component. Thus, the handle first appears as a choice of source component, not as an input label to the velocity field.

We train this source prior to fit the demonstrated action distribution. 
Motivated by OT-CFM and related coupling-based flow-matching methods~\citep{tong2024otcfm,pooladian2023multisample,liu2023rectifiedflow}, and following the observation that flow matching tends to prefer short and geometrically simple transports,
we encourage each Gaussian component to lie near the actions it is responsible for, instead of forcing all handles to start from a fixed uninformed Gaussian. We therefore fit the mixture to demonstrated actions:
\begin{equation}
    \mathcal{L}_{src}
    =
    -\mathbb{E}_{(s,a)\sim\mathcal{D}}
    \log
    \sum_{k=0}^{K-1}
    \pi_k(s)
    \mathcal{N}\!\left(a;\mu_k(s),\sigma_k^2(s)I_{d_a}\right).
    \label{eq:src-loss}
\end{equation}
This density model induces a soft responsibility for each handle:
\begin{equation}
    r_k(s,a)
    =
    \frac{
    \pi_k(s)
    \mathcal{N}\!\left(a;\mu_k(s),\sigma_k^2(s)I_{d_a}\right)
    }{
    \sum_{j=0}^{K-1}
    \pi_j(s)
    \mathcal{N}\!\left(a;\mu_j(s),\sigma_j^2(s)I_{d_a}\right)
    }.
    \label{eq:responsibility}
\end{equation}
Here $r_k(s,a)$ is a training assignment. It measures how much component $k$ is responsible for the demonstrated action under the current state. 
Unlike global source or coupling designs in computer-vision generative modeling, our assignment is conditioned on the robot state and is recomputed at every decision step. 
This distinction is important for control: the policy condition $s$ changes continuously along a closed-loop trajectory, and the same action vector can have different meanings depending on the current observation, history, and future branch.

\subsection{Orthogonal Source Lifting}

The source prior above gives us selectable components, but source selection alone does not guarantee an intervenable shared field. If two branches cross in action space, two different source components may reach the same intermediate action point with incompatible target velocities. A velocity field conditioned only on the action coordinate would then average these velocities and erase the selected source identity.

Orthogonal Source Lifting prevents this identity collapse by transporting points in a lifted space. Let
\[
    \bar{x}=[x^a,x^g]\in\mathbb{R}^{d_a+d_g},
\]
where $x^a\in\mathbb{R}^{d_a}$ is the action component and $x^g\in\mathbb{R}^{d_g}$ is an auxiliary lift coordinate. We use orthogonal anchors $\{e_k\}_{k=0}^{K-1}$, with $e_k^\top e_j=\delta_{kj}$ and typically $d_g\ge K$. Figure~\ref{fig:method-source-lift}(b) shows the fixed-anchor case for readability. In the full model, we use state-dependent lift embeddings $\tau_k(s)$ regularized toward these anchors:
\begin{equation}
    \mathcal{L}_{anc}
    =
    \mathbb{E}_{s}
    \sum_{k=0}^{K-1}
    \|\tau_k(s)-e_k\|^2.
    \label{eq:anchor-loss}
\end{equation}

For handle $k$, SL-FM lifts only the source endpoint. Given an action-space source $x^a_{0,k}$ and a demonstrated target action $a$, we define
\begin{equation}
    \bar{x}_{0,k}
    =
    [x_{0,k}^{a},\lambda\tau_k(s)],
    \qquad
    \bar{x}_1=[a,0],
    \label{eq:lift-source}
\end{equation}
where $\bar{x}_{0,k}$ is the lifted source endpoint, $\bar{x}_1$ is the lifted target endpoint, and $\lambda$ controls the lift scale.

Importantly, the target distribution is not split by $z$. Every demonstrated action is embedded into the same zero-lift plane:
\begin{equation}
    p_1(\bar{x}\mid s)
    =
    p_{\mathrm{data}}(a\mid s)\delta(x^g=0).
    \label{eq:lifted-target}
\end{equation}
Equivalently, projecting the lifted target distribution back to the action component exactly recovers the original demonstration distribution. Thus, SL-FM preserves the standard imitation target rather than replacing it with mode-conditioned targets $p(a\mid s,z)$. The handle affects only the source geometry before transport, not which target actions are included in training.

The lifted interpolation path is:
\begin{equation}
\begin{split}
    \bar{x}_{t,k} &= (1-t)\bar{x}_{0,k}+t\bar{x}_1 \\
    &= \left[
    (1-t)x^a_{0,k}+ta,\,
    (1-t)\lambda\tau_k(s)
    \right],
\end{split}
\label{eq:lifted-path}
\end{equation}
with target velocity:
\begin{equation}
    \bar{u}_{t,k}
    =
    \bar{x}_1-\bar{x}_{0,k}
    =
    \left[
    a-x^a_{0,k},\,
    -\lambda\tau_k(s)
    \right].
    \label{eq:lifted-velocity}
\end{equation}
For $t<1$, different handles remain separated in the lift coordinate even when their action coordinates overlap. At execution, only the terminal action component is applied to the robot. The lift coordinate is not a second robot action and is never sent to the environment. It is an internal integration coordinate that preserves source identity until all targets meet on the zero-lift plane. The velocity field observes the continuous lifted state $\bar{x}_t$, but it never receives a separate discrete expert label.

\subsection{Floor-Weighted Flow Training}

Given the lifted paths above, SL-FM trains one shared velocity field over all handles. 
The soft responsibility $r_k(s,a)$ from Eq.~\eqref{eq:responsibility} determines how much handle $k$ should contribute to the flow regression for the demonstrated pair $(s,a)$. 
A natural choice is to weight each lifted path by $r_k(s,a)$, so that handles closer to the demonstrated action receive larger supervision.

However, pure responsibility weighting can create dead handles when $K$ is redundant. 
In practice, some components may receive almost no assignment for many states, as illustrated by the low-responsibility handle in Figure~\ref{fig:method-source-lift}(c). 
If such handles rarely participate in flow training, their lifted trajectories can be poorly learned, leading to unreliable behavior when the evaluator switches to them at test time. 
To keep every handle trainable while preserving the responsibility structure, we use a responsibility floor:
\begin{equation}
    \gamma_k(s,a)
    =
    (1-\alpha)r_k(s,a)+\frac{\alpha}{K},
    \label{eq:floor}
\end{equation}
where $\alpha\in[0,1]$ controls the amount of uniform supervision in the objective. 
When $\alpha=0$, the method naturally reduces to pure responsibility weighting. Furthermore, larger $\alpha$ gives more training signal to low-responsibility handles. 
The floor affects only the flow-regression weights. 
Again, it does not change the deployment prior $\pi_\phi(z\mid s)$, add a policy head, or condition the velocity field on $z$.

The floor-weighted flow-matching objective is
\begin{equation}
\begin{aligned}
    \mathcal{L}_{W\text{-}FM}
    =
    \mathbb{E}_{(s,a),t,\{\epsilon_k\}}
    \sum_{k=0}^{K-1}
    \mathrm{sg}\!\left(\gamma_k(s,a)\right)
    \left\|
    v_\theta(\bar{x}_{t,k},t,s)-\bar{u}_{t,k}
    \right\|^2 ,
\end{aligned}
    \label{eq:wfm-loss}
\end{equation}
where the time step $t\sim\mathcal{U}[0,1]$ and $\mathrm{sg}(\cdot)$ stops gradients through the assignment weights. 
This prevents the flow loss from directly reshaping the posterior assignments, while still allowing the lifted source construction and the shared field to be optimized jointly during training.

Figure~\ref{fig:method-source-lift}(c) summarizes the two main training signals as source fitting and weighted flow matching. 
In the implementation, we also use scalar weights and the anchor regularizer:
\begin{equation}
    \mathcal{L}
    =
    \mathcal{L}_{W\text{-}FM}
    +
    w_{\mathrm{src}}\mathcal{L}_{src}
    +
    \beta\mathcal{L}_{anc}.
    \label{eq:total-loss}
\end{equation}
Each minibatch predicts $\{\pi_k,\mu_k,\sigma_k,\tau_k\}_{k=0}^{K-1}$ from $s$, computes $r_k$ and $\gamma_k$ from the demonstrated action, constructs all lifted paths, and updates the source prior and shared velocity field jointly.

\subsection{Deployment and source intervention}
\begin{figure}[htb]
\centering
\includegraphics[width=0.88\linewidth]{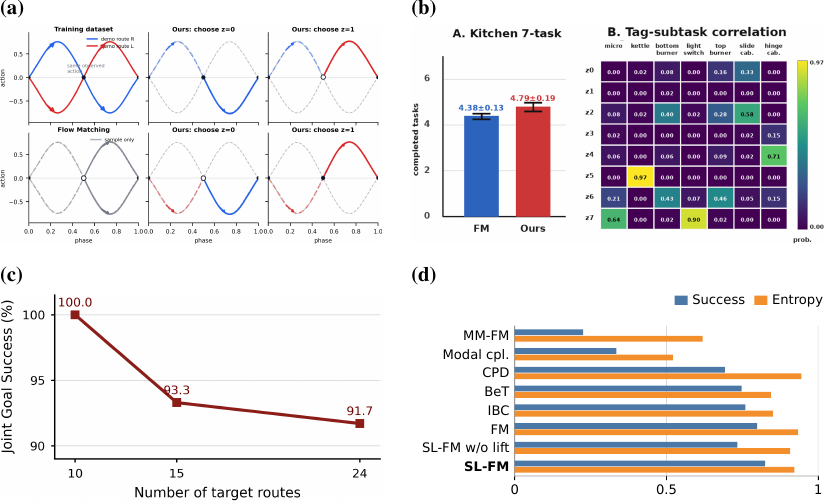}
\caption{Source intervention and empirical diagnostics for \method{}.
\textbf{(a)} Same-prefix source intervention: standard flow matching passively samples continuations, whereas \method{} exposes a local source handle $z$ for selecting valid futures through one shared velocity field.
\textbf{(b)} D4RL Kitchen: seven-task completion mean and standard deviation for FM and the $K=8$ \method{} variant (left), and source-handle--subtask correlations in rollouts (right).
\textbf{(c)} Fixed-state downstream selector results on Avoiding route targets. Joint goal success requires both matching the commanded route and successfully completing the task.
\textbf{(d)} D3IL Avoiding success and normalized route entropy across eight methods. \textit{Modal cpl.} denotes modal coupling; \textit{w/o lift} removes Orthogonal Source Lifting.}
\label{fig:combined-diagnostics}
\end{figure}

At inference time, free deployment samples \(z\sim\pi_\phi(\cdot\mid s)\): 
\[
x^a_{0,z}=\mu_z(s)+\sigma_z(s)\epsilon_z, \quad \bar{x}_{0,z}=[x^a_{0,z},\lambda\tau_z(s)],
\]
and integrates the shared field \(d\bar{x}_t/dt=v_\theta(\bar{x}_t,t,s)\).
Only the terminal action component is executed.
For intervention, the evaluator sets \(z=k\) at a chosen decision state and uses the same field. Therefore, the intervention changes only the local source endpoint, not the policy architecture or dynamics model.
Our same-prefix protocol realizes Figure~\ref{fig:combined-diagnostics}(a) by replaying an identical prefix and changing only this local handle.
The training and source-intervention pseudocode is provided in Algorithm~\ref{alg:slfm} in Appendix~\ref{sec:algorithm}.

\section{Experiments}
\suppressfloats[t]

\subsection{Protocol and Metrics}
We evaluate \method{} on D3IL Avoiding and Aligning, PushT, and D4RL Kitchen~\citep{jia2024d3il,chi2023diffusionpolicy,fu2020d4rl}.
The Avoiding task is the main source-intervention testbed because it has shared starts and frequent route decisions.
We compare against FM, source and coupling baselines (CPD, MM-FM, and Modal coupling), a direct \(z\)-conditioned field (Z-cond. field), and official BeT and IBC references.
Avoiding and Aligning report success, Avoiding also reports route entropy, PushT reports mean maximum coverage, and Kitchen reports 500-step task completion.
Additional baseline definitions, protocol details, and hyperparameters are provided in detail in the appendix.

\subsection{Task Performance Across Multimodal Benchmarks}

We first test whether the source-intervention interface preserves ordinary free-deployment imitation.
Figure~\ref{fig:combined-diagnostics}(d) shows that \method{} achieves the best Avoiding success while retaining high route entropy.
The gap between 
\textit{\method{} w/o lift}
and full \method{} indicates that learning a state-local source handle is not enough. In addition, orthogonal lifting is needed to preserve handle identity when paths cross.
MM-FM instead uses global action-space modes, which are not tied to local decision states.
Across Avoiding, Aligning, and PushT, Table~\ref{tab:performance} shows the best average score for \method{}.
The gains concentrate on tasks with more frequent local branching, while Aligning stays close to the base FM policy.
CPD, modal coupling, and direct \(z\)-conditioning can be competitive in some settings, but they lack the same state-local, source-only intervention interface.

\begin{table}[htb]
\centering
\caption{Comparison across multimodal benchmarks.
Avoid. and Align. report D3IL success. PushT reports mean maximum coverage.
Avg. is the unweighted average across the three benchmarks.}
\label{tab:performance}
\normalsize
\setlength{\tabcolsep}{8pt}
\renewcommand{\arraystretch}{1.1}
\begin{tabular}{@{}lccc|c@{}}
\toprule
Method & Avoid. & Align. & PushT & Avg. \\
\midrule
FM & 0.798 & \textbf{0.808} & 0.801 & 0.802 \\
CPD & 0.692 & 0.715 & 0.814 & 0.740 \\
MM-FM & 0.225 & 0.000 & 0.279 & 0.168 \\
Modal cpl. & 0.335 & 0.623 & 0.839 & 0.599 \\
Z-cond. field & 0.708 & 0.750 & 0.833 & 0.764 \\
\method{} & \textbf{0.825} & 0.792 & \textbf{0.852} & \textbf{0.823} \\
\bottomrule
\end{tabular}
\end{table}

We also evaluate \method{} on D4RL Kitchen, a multi-task robot manipulation benchmark~\citep{fu2020d4rl}.
The completion plot in Figure~\ref{fig:combined-diagnostics}(b) shows that the \(K=8\) \method{} variant improves over FM by \(+0.42\) completed tasks in 500-step rollouts, suggesting that the source-structured design remains useful beyond the route-control setting. The correlation heat map in Figure~\ref{fig:combined-diagnostics}(b) reports the source-handle--subtask correlation, which we discuss in the next section.

\subsection{Source-Intervention Controllability}

Task performance alone does not show that the sampled handle is an actionable cause of future behavior.
We therefore use a same-prefix counterfactual on D3IL Avoiding.
Two rollouts share an identical prefix and then differ only in the local source handle at the first high-entropy decision point in the pre-obstacle band \(y\in[-0.30,-0.02]\).
For each matched state, we replay the prefix, branch into all six unordered pairs of \(z\in\{0,1,2,3\}\), force each selected handle for five consecutive policy calls, and then return both branches to free source resampling.

Table~\ref{tab:decision-intervention} shows that changing the source handle redirects the future while preserving task execution.
\method{} changes the final route in \(91.1\%\) of paired interventions and achieves a \(57.8\%\) both-success route-change rate.
This last metric is stricter than route change alone: it counts only pairs where both branches succeed and nevertheless take different routes.
FM noise resampling and direct \(z\)-conditioning can also change routes, but they produce fewer successful route changes.
The same-handle control produces no route changes, confirming that the effect comes from changing the source handle rather than replay instability.

\begin{table}[htb]
\centering
\caption{Same-prefix source intervention on D3IL Avoiding. From an identical pre-decision prefix, paired rollouts differ only in the five-policy-call local intervention. Outside the window, stochastic draws are matched and the policies resume free sampling. \(z\)-field is a \(z\)-conditioned velocity field.}
\label{tab:decision-intervention}
\normalsize
\setlength{\tabcolsep}{4pt}
\begin{tabular}{llcccc}
\toprule
Method & Intervention & Future & Route & Both & Both-succ. \\
 & & sep. & changed & success & route chg. \\
\midrule
FM & noise resample & 0.190 m & 0.767 & 0.428 & 0.286 \\
\(z\)-field & change \(z\) input & 0.273 m & \textbf{0.930} & 0.455 & 0.421 \\
\method{} & same handle & 0.000 m & 0.000 & 0.787 & 0.000 \\
\method{} & change handle & \textbf{0.287 m} & 0.911 & \textbf{0.639} & \textbf{0.578} \\
\bottomrule
\end{tabular}
\end{table}

Appendix~\ref{sec:free-rollout-diagnostics} provides the detailed analysis of source-handle--route mutual information under free resampling in Figure~\ref{fig:free-rollout-mi}.

In Kitchen, the correlation heat map in Figure~\ref{fig:combined-diagnostics}(b) shows a non-uniform association between source handles and subtasks.
The strongest entries align specific handles with subtasks: \(z5\) with kettle, \(z7\) with microwave and light switch, \(z4\) with hinge cabinet, and \(z2\) with slide cabinet and bottom burner. Notably, slide cabinet and bottom burner are spatially and procedurally related subtasks, and their shared association with \(z2\) suggests that the source handles can capture semantically meaningful structure rather than only arbitrary rollout variation.
Together, these results suggest that the handles capture subtask-specific progress patterns in long-horizon manipulation.

\subsection{Downstream Route Control}

We next test whether a downstream module can use the exposed handle.
With the low-level \method{} policy frozen, a high-level selector chooses \(z\) every five consecutive policy calls from a fixed initial state.
Free source resampling still solves the task, but it achieves 0\% commanded-route success for the specified targets.
CEM-retrieved source schedules and a PPO selector initialized from CEM elites reach \(91.7\%\) joint route-and-task success over all 24 route targets~\citep{rubinstein1999cem,schulman2017ppo}.
Figure~\ref{fig:combined-diagnostics}(c) shows the same trend under increasingly broad route sets: the selector reaches \(100.0\%\) joint success over 10 target routes, \(93.3\%\) over 15 routes, and \(91.7\%\) over all 24 routes.
This result demonstrates that the source handle can serve as an action space for high-level control, analogous to skill-level decision abstractions, while the same-prefix experiment remains the primary causal evidence.

Additional design ablations and sensitivity sweeps are reported in Appendix~\ref{sec:design-ablations}.

\section{Conclusion}

We presented \methodfull{}, a flow-matching policy that exposes an intervenable source interface without conditioning the velocity field on a discrete code.
The central mechanism is \mech{}: sources are lifted into orthogonal coordinates while every target remains on the zero-lift plane.
This preserves the original conditional data distribution at \(x^g=0\), but makes source choices externally addressable during deployment.
Across crossing-flow, D3IL, PushT, and Kitchen diagnostics, \method{} remains competitive in free deployment and redirects behavior under same-prefix source interventions.

\clearpage
\section*{AI Use Statement}

The authors used generative AI tools to assist with language polishing, the organization and presentation of experimental data and figures, preliminary literature exploration, and feedback on manuscript drafts. All AI-assisted outputs, including literature-related suggestions and numerical or visual material, were reviewed and verified by the authors. The authors take full responsibility for the study's ideas, methodology, experiments, analysis, claims, and final text.

\bibliographystyle{iclr2027_conference}
\bibliography{refs}

\clearpage
\appendix
\section{Training and Source Intervention}
\label{sec:algorithm}

\begin{center}
\centering
\refstepcounter{algorithm}\label{alg:slfm}
\begin{minipage}{0.96\linewidth}
\normalsize
\hrule height 0.8pt
\vspace{1pt}
\textbf{Algorithm \thealgorithm} \method{} Training and Source Intervention\par
\vspace{1pt}
\hrule
\vspace{2pt}
\textbf{Input:} demonstrations \(\{(s_i,a_i)\}\), handles \(K\), anchors \(\{e_k\}\), lift scale \(\lambda\), floor \(\alpha\), weights \(w_{\mathrm{src}},\beta\).\par
\textbf{Output:} source prior \(p_\phi\), shared field \(v_\theta\), and source-handle rollout interface.\par
\vspace{2pt}
\begin{tabular}{@{}r@{\hspace{0.45em}}p{0.86\linewidth}@{}}
1: & \textbf{for} minibatch \((s,a)\) \textbf{do}\\
2: & Predict \(\{\pi_k,\mu_k,\sigma_k,\tau_k\}_{k=0}^{K-1}\) from \(s\).\\
3: & Compute \(r_k(s,a)\) and \(\gamma_k=(1-\alpha)r_k+\alpha/K\).\\
4: & \textbf{for} handle \(k=0,\ldots,K-1\) \textbf{do}\\
5: & \(\quad\)Sample \(\epsilon_k\sim\mathcal{N}(0,I_{d_a})\) and set \(x^a_{0,k}=\mu_k(s)+\sigma_k(s)\epsilon_k\).\\
6: & \(\quad\)Lift source and target: \(\bar{x}_{0,k}=[x^a_{0,k},\lambda\tau_k(s)]\), \(\bar{x}_1=[a,0]\).\\
7: & \(\quad\)Sample \(t\sim\mathcal{U}[0,1]\); compute \(\bar{x}_{t,k}\), \(\bar{u}_{t,k}\).\\
8: & \textbf{end for}\\
9: & Update \((\theta,\phi)\) with \(\mathcal{L}_{W\text{-}FM}+w_{\mathrm{src}}\mathcal{L}_{src}+\beta\mathcal{L}_{anc}\).\\
10: & \textbf{end for}\\
11: & \textbf{Free rollout:} sample \(z\sim\pi_\phi(\cdot\mid s)\), sample \(\bar{x}_{0,z}\), integrate \(v_\theta(\bar{x}_t,t,s)\), and execute the action component.\\
12: & \textbf{Intervention:} set \(z=k\), sample \(\bar{x}_{0,k}\), integrate the same field, and execute only the action component.\\
\end{tabular}
\vspace{2pt}
\hrule
\end{minipage}
\end{center}

\section{Experimental Details}

\paragraph{Baselines and ablations.}
Unless labeled as an ablation, \method{} uses source-only \(z\), \mech{}, one latent-free shared velocity field, and the responsibility floor.
\begin{description}
\item[FM.]
The standard conditional flow-matching policy uses a single Gaussian source and one field; its intervention baseline is source-noise resampling rather than handle selection.

\item[CPD.]
Conditional Prior Distribution replaces the fixed source with a state-conditioned single Gaussian trained toward demonstrated actions, but has no \(K\)-way handle, responsibility assignment, or lifted tag coordinate.

\item[MM-FM.]
MM-FM uses a global action-space GMM source; its component index is a global action cluster rather than a state-local decision handle.

\item[Modal coupling.]
Modal coupling uses fixed global source anchors and hard action-cluster/source-target pairings, so its pairings are not recomputed as state-local decisions.

\item[\(z\)-conditioned field.]
This direct-conditioning baseline feeds the discrete handle into \(v_\theta(x_t,t,s,z)\), allowing mode-specific dynamics; \method{} keeps the field shared and lets \(z\) affect only the source endpoint.

\item[BeT.]
Behavior Transformer is included as an official D3IL Avoiding reference and represents multimodality through discrete behavior/action tokens rather than flow source geometry.

\item[IBC.]
Implicit Behavioral Cloning is another official Avoiding reference; it scores candidate actions with an implicit model and has no flow source or source-level intervention variable.

\item[\method{} w/o lift.]
This ablation keeps the state-conditioned source mixture and source-only handle but removes the auxiliary lift coordinate, testing whether a learned local handle alone preserves identity through crossings.
\end{description}

\paragraph{Evaluation protocol.}
Avoiding follows the official D3IL state benchmark protocol and reports success and normalized route entropy over the 480-trajectory, three-seed evaluation. PushT reports mean maximum coverage, and Kitchen reports 500-step task completion with eight environments and three evaluation seed offsets.

\paragraph{Representative defaults.}
A representative D3IL Avoiding run uses \(K=4\), \(d_g=K\), \(\sigma_{\mathrm{iso}}=1.0\), responsibility floor \(\alpha=0.25\), anchor/balance weights \(0.3/0.01\), 10 Euler steps, and 2000 training epochs.
The encoder, shared field, and source heads are residual MLPs with six 256-wide Mish layers; other domains keep the same interface while adjusting task-specific backbones or \(K\).

\section{Design Ablations and Sensitivity}
\label{sec:design-ablations}

\begin{figure}[H]
\centering
\includegraphics[width=\linewidth]{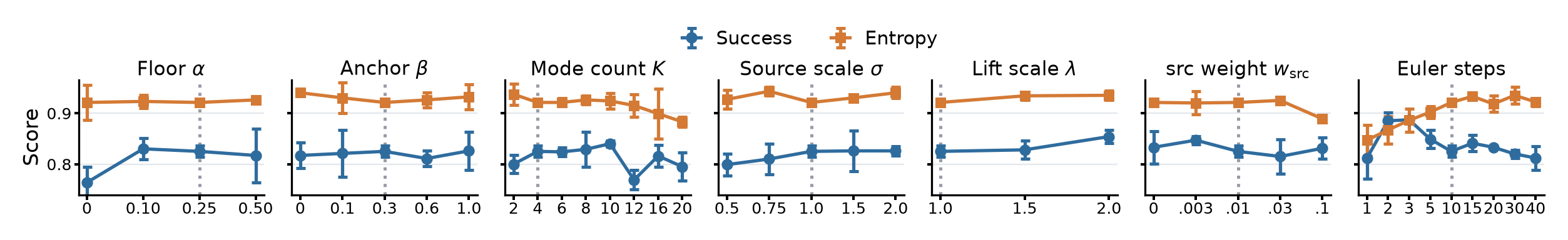}
\caption{Hyperparameter and deployment-cost sensitivity on D3IL Avoiding. Each plot varies one setting while keeping the remaining defaults fixed. Blue curves report success and orange curves report normalized route entropy. All plotted points use the official 480-trajectory, three-seed evaluation protocol.}
\label{fig:param-sensitivity}
\end{figure}

Figure~\ref{fig:param-sensitivity} summarizes dense Avoiding sensitivity sweeps.
The responsibility floor only changes the flow-regression weights \(\gamma_k\).
Moderate floors improve success while preserving route entropy, supporting their role as dead-handle robustness rather than collapse-inducing regularization.
Nearby anchor penalties, source scales, source-fitting weights, and mode counts \(K=4\)--\(10\) remain viable.
Very large \(K\) is less reliable, likely because a fixed minibatch provides noisier responsibility-weighted updates for each handle.
Unstable lift scales and excessive Euler steps also reduce robustness.
All plotted Avoiding points use the official 480-trajectory, three-seed protocol.

\FloatBarrier

\section{Free-Rollout Source--Route Diagnostics}
\label{sec:free-rollout-diagnostics}

\begin{figure}[H]
\centering
\includegraphics[width=0.78\linewidth]{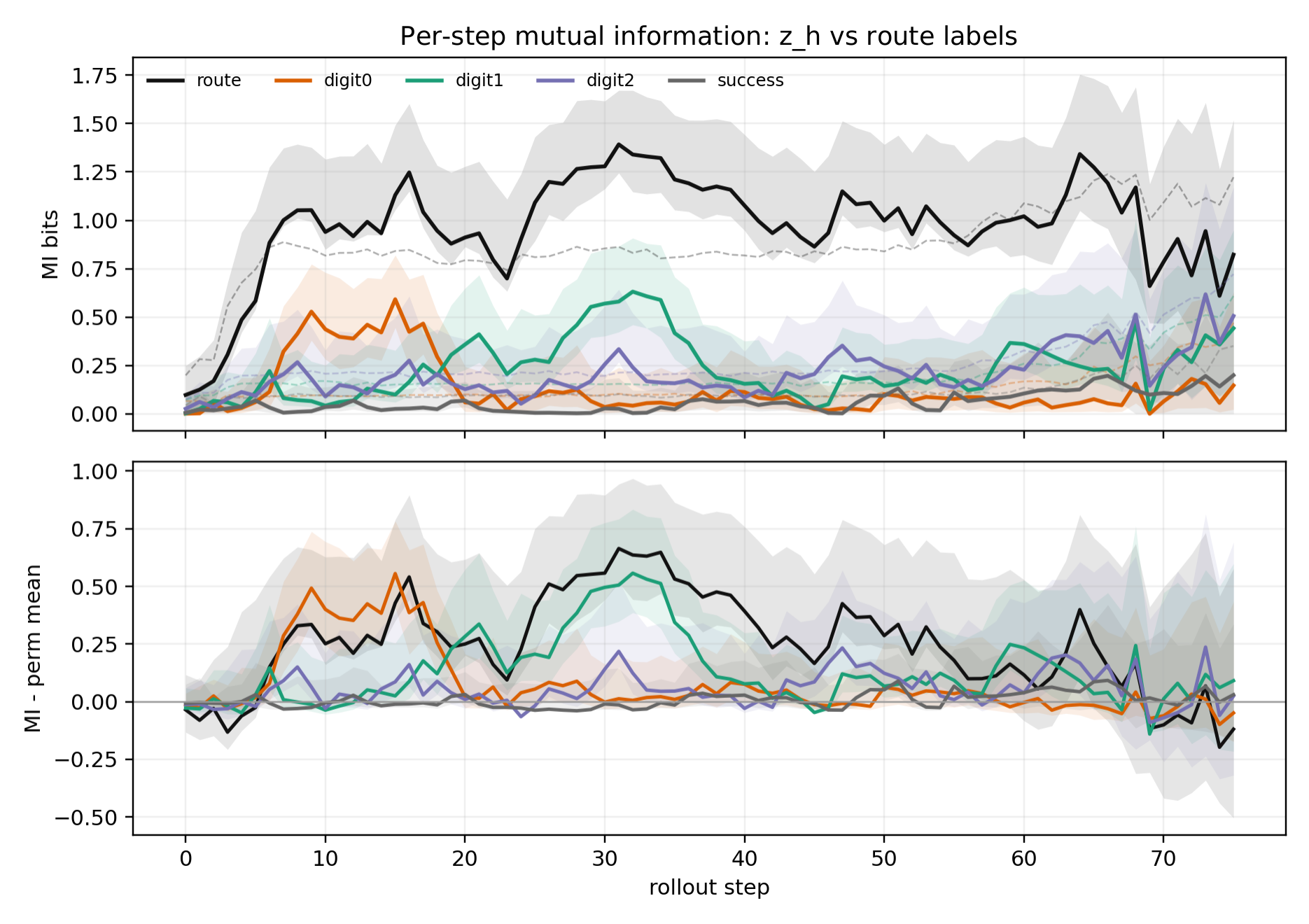}
\caption{Free-rollout source--route diagnostics on D3IL Avoiding. Per-step mutual information between sampled source handles and realized route labels under free source resampling, through step 75 with bootstrap 95\% confidence intervals; peaks align with route decision phases. The upper panel reports mutual information in bits; the lower panel subtracts the permutation mean. Route digits digit0, digit1, and digit2 denote the branch choices around the first, second, and third obstacle regions, respectively.}
\label{fig:free-rollout-mi}
\end{figure}

We complement the same-prefix intervention results in Table~\ref{tab:decision-intervention} with free-rollout diagnostics on D3IL Avoiding.
In Avoiding, each successful trajectory can be encoded by three route digits, where digit0, digit1, and digit2 denote the branch choices around the first, second, and third obstacle regions, respectively.
Figure~\ref{fig:free-rollout-mi} reports the mutual information between sampled handles and these realized route digits over time.
The digit0 curve reaches its peak before the first obstacle, showing that early handle samples are most informative about the first branch choice.
The digit1 curve peaks later, before the second obstacle, consistent with the second route decision occurring after the first branch has been resolved.
The digit2 curve is weaker: near the final obstacle, the current position already strongly constrains the remaining route, so \(z\) carries less additional information under free rollout.
Nevertheless, the intervention results in Table~\ref{tab:decision-intervention} show that changing \(z\) can still force meaningful future changes.
Mutual information with success remains small throughout, indicating that the handles track branch identity rather than merely success or failure.

\section{Additional Discussion}

\paragraph{Planning and reinforcement learning.}
The exposed source handle is an interface rather than a complete planner.
It can be combined with MPC or RL by letting a planner, value function, or high-level policy choose or constrain \(z\), while the flow policy executes low-level continuous actions.

\paragraph{Semantic conditioning.}
The handle can also be grounded by semantic signals.
Text instructions, goal images, or vision-language embeddings could bias \(\pi_\phi(z\mid s)\) or select interventions, connecting high-level intent to continuous control without conditioning the shared velocity field directly on a discrete mode.

\paragraph{Large heterogeneous imitation data.}
Because responsibilities are learned without manual mode labels, source handles are promising for large imitation corpora that mix styles, imbalanced modes, and non-expert or mixed-quality trajectories.
In this setting, the source prior can provide a compact interface for separating reusable behavior patterns while retaining a single shared action generator.

\section{Additional Kitchen Results}

Table~\ref{tab:kitchen500} gives the full 500-step Kitchen diagnostic behind the completion plot in Figure~\ref{fig:combined-diagnostics}(b), using eight environments and seed offsets \(0,8,16\).
Here \(\alpha\) is the responsibility floor, \(\beta\) the anchor weight, \(w_{\mathrm{src}}\) the source-fitting weight, and \(\lambda\) the lift scale.
Four of five source-structured variants exceed FM on 7-task completion, and the variant with \(w_{\mathrm{src}}=0.005\) gives the best four-target score.

\begin{table}[htbp]
\centering
\caption{D4RL Kitchen 500-step diagnostic at checkpoint 500. Four-target scores sum microwave, kettle, bottom burner, and light switch. The 7-task scores are reported as mean ± standard deviation.}
\label{tab:kitchen500}
\normalsize
\setlength{\tabcolsep}{3.0pt}
\begin{tabular}{lcc}
\toprule
Method / variant & 7-task mean & 4-target mean \\
\midrule
\method{} \(\alpha{=}0.10,\beta{=}0.3,\lambda{=}10\) & \textbf{4.79 \(\pm\) 0.19} & 2.71 \\
\method{} \(\alpha{=}0.25,\beta{=}0.3,w_{\mathrm{src}}{=}0.005,\lambda{=}10\) & 4.71 \(\pm\) 0.14 & \textbf{2.96} \\
\method{} \(\alpha{=}0.25,\beta{=}0.3,\lambda{=}10\) & 4.67 \(\pm\) 0.80 & 2.83 \\
\method{} \(\alpha{=}0.10,\beta{=}0.5,\lambda{=}15\) & 4.62 \(\pm\) 0.22 & 2.62 \\
\method{} \(\alpha{=}0.25,\beta{=}0.5,\lambda{=}10\) & 4.29 \(\pm\) 0.26 & 2.54 \\
FM & 4.38 \(\pm\) 0.13 & 2.54 \\
\bottomrule
\end{tabular}
\end{table}

\FloatBarrier
\section{Additional Qualitative Results}

\begin{figure}[htb]
\centering
\includegraphics[width=0.88\textwidth]{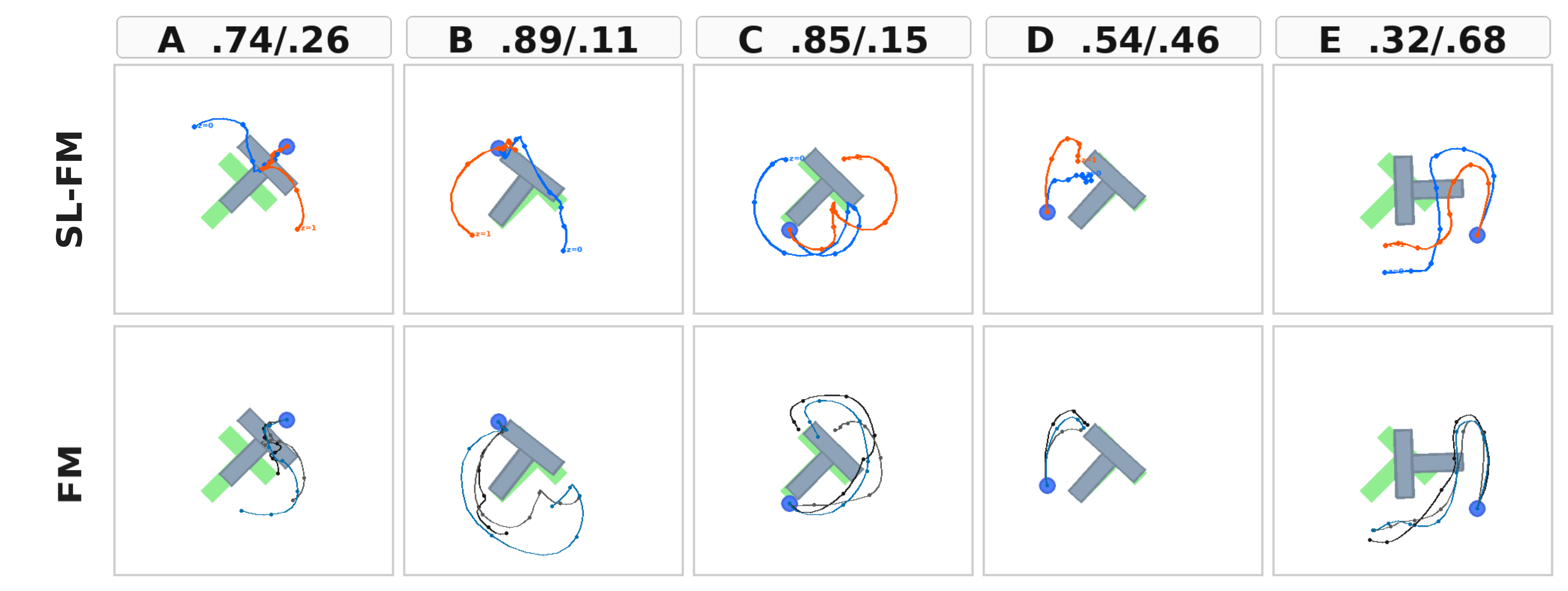}
\caption{Additional PushT branching examples. Columns show selected PushT states. Forced source choices in the source-structured policy expose visibly separated contact strategies, whereas FM source-noise resampling provides stochastic variation without an explicit persistent handle.}
\label{fig:pusht-branch-appendix}
\end{figure}

Figure~\ref{fig:pusht-branch-appendix} visualizes PushT branching at matched states.
Forced source choices in the source-structured policy produce distinct contact strategies around the T block, while FM source-noise resampling gives stochastic variation without a persistent index for selecting one strategy.
These examples complement the aggregate PushT scores by showing that the exposed handle corresponds to qualitatively different manipulation choices.

\begin{figure}[htb]
\centering
\includegraphics[width=0.88\textwidth]{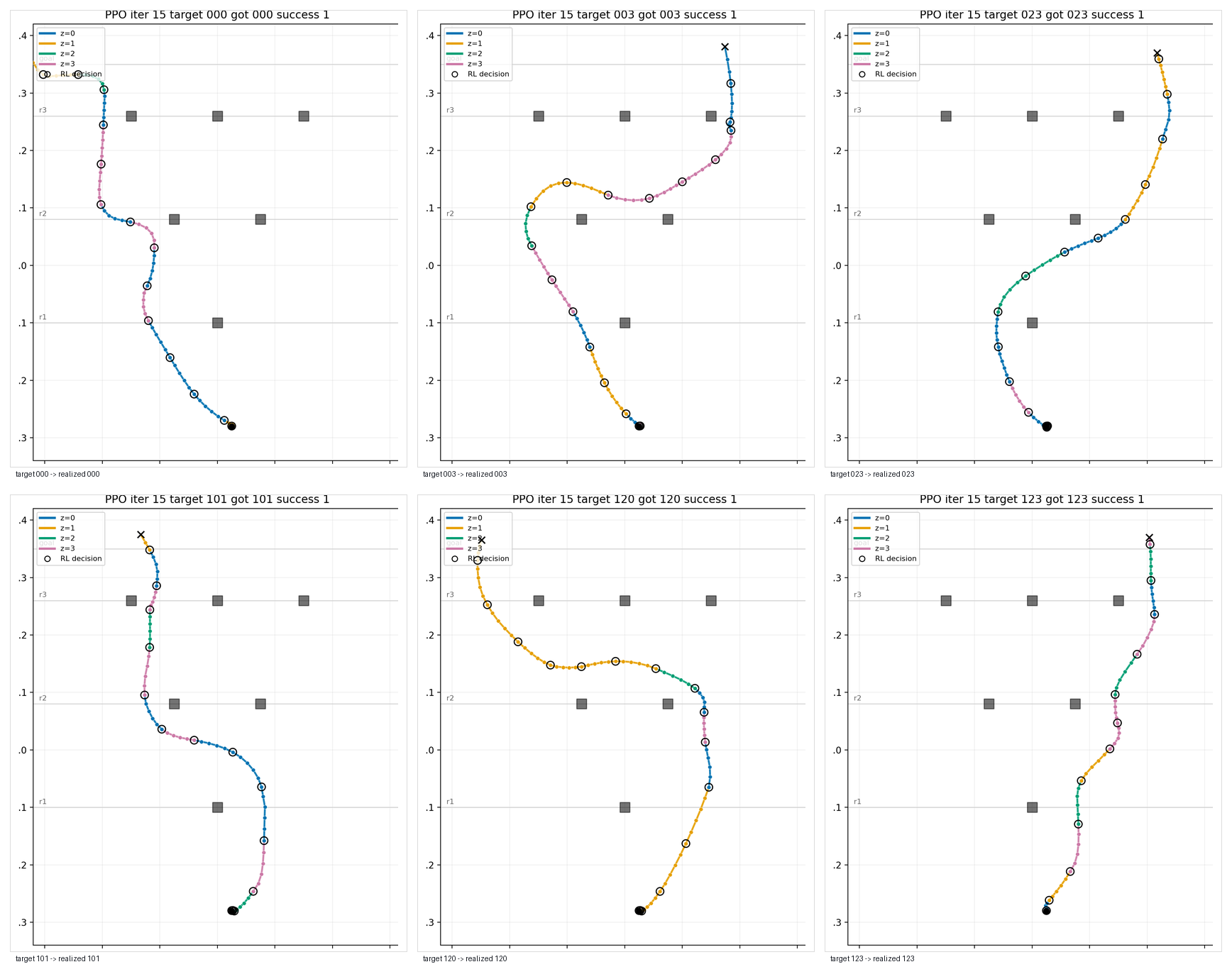}
\caption{Representative successful downstream route-control trajectories on Avoiding. The low-level \method{} policy is frozen, while a high-level selector chooses one source handle every five policy calls. The shown fixed-start PPO checkpoint reaches \(22/24\) joint route-and-task success; colors indicate the executed source handle and black circles mark selector decisions.}
\label{fig:avoiding-downstream-trajs}
\end{figure}

Figure~\ref{fig:avoiding-downstream-trajs} shows successful Avoiding trajectories from the downstream selector.
Black circles mark selector decisions and colors indicate the active source handle.
The trajectories reach different commanded routes from the same start, illustrating that a high-level module can use the handle as a compact route-control interface after the low-level \method{} policy is frozen.

\end{document}